\documentclass[journal]{IEEEtran}  

\usepackage{multirow}
\usepackage[pdftex]{graphicx}
\usepackage{cite}
\usepackage{mdwmath} 
\usepackage{mdwtab}
\usepackage{eqparbox}
\usepackage{mathtools}
\usepackage[utf8]{inputenc}
\usepackage{multicol}
\usepackage[]{subcaption}  
\usepackage{amsmath}  
\usepackage{caption}

\usepackage{dsfont}
\usepackage{bm}
\usepackage{physics}  
\usepackage{amssymb}  
\usepackage{csvsimple}  
\usepackage{algorithm}
\usepackage{algpseudocode}
\usepackage{listings} 
\usepackage{amsfonts}

\usepackage{amsthm} 
\usepackage{gensymb} 
\usepackage{mathtools}
\usepackage{hyperref} 
\hypersetup{
  colorlinks   = true,    
  urlcolor     = blue,    
  linkcolor    = blue,    
  citecolor    = red      
}

\newcommand*{\nline}{\vspace{0.5mm}\noindent}

\DeclareMathOperator{\E}{\mathbb{E}}

\newcommand{\argmin}{\arg\,\min}
\renewcommand{\pb}[1]{\left({#1}\right)} 

\newcommand{\sqb}[1]{\left[{#1}\right]} 
\newcommand{\nn}{\nonumber} 
\renewcommand{\norm}[1]{\left\lVert{#1}\right\rVert}
 
\newcommand{\mbf}[1]{\mathbf{#1}}

\allowdisplaybreaks
\begin{document}

\title{Sparse Signal Models for Data Augmentation in Deep Learning ATR}
\author{Tushar Agarwal, Nithin Sugavanam
        and~Emre Ertin

\thanks{T. Agarwal, Nithin Sugavanam, and E. Ertin are with the Department
of Electrical and Computer Engineering, The Ohio State University, Columbus,
OH, 43210 USA}
\thanks{Corresponding Author: T. Agarwal (agarwal.270@buckeyemail.osu.edu)}
\thanks{Code available at \url{https://github.com/SENSE-Lab-OSU/mstar_data_aug}}
\thanks{This work has been submitted to the IEEE for possible publication. Copyright may be transferred without notice, after which this version may no longer be accessible.}
}

\maketitle 


\begin{abstract}
Automatic Target Recognition (ATR) algorithms aim to classify a given Synthetic
Aperture Radar (SAR) image into one of the known target classes using
information gleaned from a set of training images available for each class.
Recently, deep learning methods have been shown to achieve state-of-the-art
classification accuracy if abundant training data is available, sampled
uniformly over the classes, and their poses. In this paper, we consider the ATR
problem when a limited set of training images are available. We propose a data
augmentation approach to incorporate SAR domain knowledge and improve the
generalization power of a data-intensive learning algorithm, such as a
Convolutional neural network (CNN). The proposed data augmentation method
employs a physics-inspired limited persistence sparse modeling approach,
capitalizing on commonly observed characteristics of wide-angle synthetic
aperture radar (SAR) imagery. Specifically, we fit over-parametrized models of
scattering to limited training data and use the estimated models to synthesize
new images at poses and sub-pixel translations not available in the given data
to augment the limited training data. We exploit the sparsity of the scattering
centers in the spatial domain and the smoothly-varying structure of the
scattering coefficients in the azimuthal domain to solve the ill-posed problem
of the over-parametrized model fitting.  The experimental results show that for the
training data starved region, the proposed method provides significant gains in
the resulting ATR algorithm's generalization performance.
\end{abstract}

\IEEEpeerreviewmaketitle

\section{Introduction}\label{intro}
Synthetic aperture radar (SAR) sensors provides day and night  high-resolution imaging capability robust to weather and other environmental factors. The SAR sensor consists of a moving radar platform with a collocated receiver and transmitter that traverses a wide aperture in azimuth, acquiring coherent measurements of scene reflectivity.  Returns for multiple pulses across the synthesized aperture are combined and coherently processed to produce high-resolution SAR imagery. SAR imaging system achieves a high spatial resolution in both the radial direction, termed as range, as well as the orthogonal direction, termed as cross-range. The range resolution is a function of the bandwidth of the signal used in illumination.  The cross-range resolution is a function of the antenna aperture's size and the persistence of scattering centers~\cite{moses2004wide}. A significant fraction of the energy in the back-scattered signal from the scene is due to a small set of dominant scattering centers resolved by the SAR sensor. The localization of back-scatter energy provides a distinct description of targets of interest~\cite{Potter_attributed_scattering_1997}, for example in the case of  man-made objects such as civilian and military vehicles. This sparsity structure has been utilized in \cite{cetin_2014_SAR,potter_SAR_survey_2010} to design features like peak locations and edges. that succinctly represents the scene.
In early work, these hand-crafted features are used in solving the target recognition problem in a statistical framework. Notably, the template-based methods exploit the geometric structure and variability of these features in the scattering centers in~\cite{templateMatching_potter:2010,template_matching_potter:2011} to distinguish between the different target categories. The target signature of each of the scattering center varies with the viewing angle of the sensor platform. Statistical methods can explicitly model and utilize this low-dimensional manifold structure of the scattering center descriptors \cite{abdelrahman2015mixture,Numax_embedding_classification:2015} for improved decisions as well as integrating information across views ~\cite{sequentialSAR_Ertin_2016,HMM_radar_ATR:2008}.

However, ATR algorithms based on these hand-crafted features are limited to the information present in these descriptors, and lack the generalization ability with respect to variability in clutter, pose, and noise. With the advent of data-driven algorithms such as artificial neural networks (ANN)~\cite{Lecun_NN:1998}, an appropriate feature set and a discriminating function can be jointly estimated using a unified objective-function. Recent advances in techniques to incorporate deep hierarchical structures used in ANN~\cite{KrizhevskyImageNet:2017,Hinton504} has led to the widespread use of these methods to solve inference problems in a diverse set of application areas. Convolutional Neural Networks (CNN), in particular, have been used as an automatic feature extractors for image data. These methods have also been adopted in solving the ATR problem using SAR images~\cite{CNN_SAR_ATR:2016}. There have been several efforts in this direction, including the state-of-the-art ATR results on the MSTAR data-set in \cite{zhong2017enlightening}.  These results establish that the CNN could be effective in radar image classification and provided sufficient training data.  However, this approach of designing ATR algorithms for new sensors operating in different bands and elevations, with limited training data from targets of interest is not feasible, as the scattering behavior changes substantially as the wavelength of the operation changes. The major challenge is that Neural Networks usually require large data-sets to have good generalization performance. In general, labeled radar image data is not readily available in abundance, unlike other image data-sets. In this paper, we address the scarcity of training data and provide a general method that utilizes a model-based approach to capture and exploit the underlying scattering phenomenon to enrich the training data-set.

 Transfer learning is one of the most effective techniques to handle the availability of limited training data. Transfer-learning uses the model parameters,  estimated using a similar data-set such as Image-net~\cite{imageNet_2015}, as initialization for solving the problem of interest, typically using convolutional neural networks with little to no fine-tuning. There have been numerous experiments supporting the benefits of transfer learning, including two seminal papers~\cite{yosinski2014transferable} and \cite{oquab2014learning}. However, radar images are significantly different from regular optical images. In particular, SAR works in the wavelength of $1 cm.$ to $10 m.$ while visible light has a wavelength of the order of $1 nm.$. As a result, most surfaces in natural scenes are rough at visible wavelengths, leading to diffused reflections.
In contrast, microwaves from radar transmitters undergo specular reflections. This difference in scattering behavior leads to substantially different images in SAR and optical imaging. Since specular reflections dominate the scattering phenomenon, the images are sensitive to instantaneous factors like the imaging device's orientation and background clutter. Therefore, readily available optical-imagery based deep neural network models like Alex-net and VGG16~\cite{simonyan2014deep} are not suitable for transferring knowledge to the SAR domain. In this paper we pursue an alternative strategy of data augmentation of limited datasets using a principled approach that exploits the phenomenology of the RF backscatter data.

The paper is structured as follows. We first review relevant research work and outline our contributions in sections \ref{RW} and \ref{contri}, respectively. In sections \ref{data} and \ref{network}, we describe the data-set and network architecture in detail. In section \ref{teach}, we provide an overview of our strategy and then describe the details of our pose-synthesis methodology in section \ref{pose}. Next, we present the details of the experiments and corresponding results in sections \ref{Exp} and \ref{Res}, respectively, which provide the empirical evidence for the effectiveness of the proposed data augmentation method. We conclude with some possible directions for future research in section \ref{Conc}.

\subsection{Related Work}\label{RW}
Over-fitting is a modeling error common to data-driven machine learning methods when the learned classifier function is too closely aligned to the training data points and therefore fails to generalize to data points outside the support of the training set. Over-fitting problem is exacerbated with smaller training sets. Several methods have been proposed to reduce over-fitting and improve generalization performance. Typically, the ill-posed problem of fitting an overparametrized function to data is solved by using regularizers that impose structure and constraints in the solution space. The norm of the model parameters serves as a standard regularizing function. This keeps the parameter values small with 2-norm ($||\cdot||_2$ called $L_2$ loss) or sparse with 1-norm($||\cdot||_1$ called $L_1$ loss). Furthermore, the optimization algorithms, such as stochastic gradient descent and mirror descent, implicitly induce regularization~\cite{gunasekarBias18a,iclr_2019_implicit_reg}. Dropout, introduced by Srivastava et al. \cite{srivastava2014dropout}, is another popular method, specifically for deep neural networks. The idea is to randomly switch off certain neurons in the network by multiplying a Bernoulli random variable with a predefined probability distribution. The overall model learned is an average of these sub-models, giving improved generalization performance. Batch Normalization is another way to improve generalization performance proposed by Ioffe and Szegedy \cite{ioffe2015batch}. They proposed normalizing all neuron values of designated layers continuously while training along with an adaptive mean and variance that also get learned as part of the back-propagation training regime. Finally, recent work~\cite{iclr_2019_overParam} has established the benefit of over-parameterizing in implicitly regularizing the optimization problem and improving the generalization performance.

Transfer Learning is another approach for improving generalization performance in the limited availability of data. Pan and Yang \cite{pan2009survey} provide a comprehensive overview illustrating the different applications and performance gains of transfer learning.  For radar data, Huang et al. \cite{huang2017transfer} suggested a promising approach along this direction by using a large corpus of SAR data to train feature-extractors in an unsupervised manner.

For SAR data, there exist several neural network architecture based approaches to improve generalization. Chen et al. \cite{chen2016target} restrict the effective degrees of freedom of the network by using a fully convolutional network. Lin et al. \cite{lin2017deep} propose a convolutional highway network to tackle the problem of limited data availability. In \cite{fu2018small}, authors design a specialized ResNet architecture that learns effectively even when training data-set is small.

Few other approaches focus on learning a special type of feature. Dong et al. \cite{dong2014sparse} generate an augmented monogenic feature vector followed by a sparse representation-based classification. In \cite{song2016sar}, authors use hand-designed features with supervised discriminative dictionary learning, to perform SAR ATR. Song et al. used a Sparse Representation based Classification (SRC) approach in \cite{song2016sparse}. In \cite{huang2018sar}, Huang et al. designed a joint low rank and sparse dictionary to denoise the radar image while keeping the main texture of the targets. Much recently, Yu et al. \cite{yu2019high} proposed a combination of Gabor features and features extracted by neural networks, for better classification performance.

Data Augmentation is another regularization strategy to reduce the generalization error while not affecting the training error~\cite{augment_ICLR_2017,Goodfellow-et-al-2016}, and is the main focus of this paper. The main idea is to use domain-specific transformations to augment the original training data-set. J. Ding et al. \cite{ding2016convolutional} explored the effectiveness of conventional transformations used for optical images, viz. translations, noise addition, and linear interpolation (for pose synthesis). They report marginal improvements in classification performance on the MSTAR data-set. More recently, \cite{marmanis2017artificial} proposed a Generative Adversarial Network (GAN) to generate synthetic samples for the augmentation of SAR data but did not report any significant improvements in the error rate of the ATR task. In another effort using a GAN, Gao et al. \cite{gao2018deep} used two jointly trained discriminators with a non-conventional architecture. They further used the trained generator to augment the base data-set and reported significant improvements. Cha et al. \cite{cha2018improving} use images from a SAR data simulator and refine them using a learned function from real images. Simple rotations of radar images were considered as a data-augmentation method \cite{zhai2019sar}. In \cite{zhong2017enlightening}, Zhong et al. suggested key ideas to incorporate prior knowledge in training the model. They added samples flipped in the cross-range dimension with a reversed sign of the azimuthal angle. Such flip-augmentation exploits the symmetric nature of most objects in the MSTAR Dataset. They also added a loss related to a secondary objective to the primary objective of classification. The authors used the pose prediction (azimuthal angle) as the secondary objective of the network. They empirically showed that this helps by adding meaningful constraints to the network learning. Thus, the network is more informed about the auxiliary confounding factor, improving its generalization capability. There have been several other models proposed to tackle the ATR problem with the MSTAR data-set, including \cite{alver2018sar} and \cite{wang2017joint}, but we primarily build upon Zhong et al.'s \cite{zhong2017enlightening} work.

\subsection{Contributions}\label{contri}

In this work, we introduce a novel data-augmentation method for SAR domain, following a principled approach that exploits the phenomenology of the RF backscatter data over the azimuth and frequency domains. This paper is an extension of our previous work \cite{agarwal2020sparse} with additional results that include a comparison to other existing techniques and an ablation study of the components of the proposed technique.

First we introduce an approach for pose synthesis that models and exploits the limited persistence of scatterers over the azimuth domain. Specifically we first transform the image into the polar frequency domain to obtain the samples in the phase history domain. We then construct a model motivated by the scattering behavior of canonical reflectors in this phase history domain. This model captures the phenomenology of viewing-angle dependent anisotropic scattering behavior of man made-objects, and provides realistic imagery at poses outside the training dataset, with quality far surpassing previous approaches such as linearly interpolation in image domain~\cite{ding2016convolutional}.

Second, with modeling in the complex valued phase history domain, our algorithm can create realistic sub-pixel shift augmentations capturing the well known scintillation effects in SAR imagery. These sub-pixel shifts are not possible in the traditional image domain using standard interpolators (linear, cubic, etc) as the complex valued interpolation kernel need to be appropriately designed taking into account azimuth and frequency window of the sensor. We hypothesize that these two factors are essential to improving the network's knowledge about the SAR imaging systems' underlying physics. 

Third, we focus on a state of the art deep learning classifier for SAR ATR~\cite{zhong2017enlightening} and the MSTAR dataset, and provide extensive simulation studies to illustrate the learning performance for different training dataset sizes with un-augmented and augmented approaches to training. Our results show a significant boost in generalization performance over both un-augmented and augmentation with previously suggested approaches. In particular, for the MSTAR dataset when the training data-set is reduced by a factor of 32, the proposed augmentation algorithm reduces the test error by more than 42\% as compared to the baseline approach that includes image domain flips and integer pixel translations.

It is important to note that our data-augmentation based strategy is generic and decoupled with the network architectures proposed in other works like \cite{lin2017deep}, \cite{chen2016target}. Therefore, the proposed augmentation strategy may yield even further improvements in conjunction with the methods mentioned above. Our objective here is to demonstrate the benefits of the proposed data-augmentation strategy. Hence, apart from data-augmentation, we only use Zhong et al.'s \cite{zhong2017enlightening} multi-task learning paradigm.

\section{Model Based SAR Data Augmentation}\label{DA}
An approach to ATR algorithm design is  to learn a parametric Neural Network
classifier $g$, with parameters $w \in \mathbb{R}^{d_w}$, that predicts an
estimate of output labels, $Y \in \mathbb{R}^{d_Y}$ for an input $X \in
\mathbb{C}^{d_X}$ , i.e. $\hat{Y} = g(X;w)$ where $d_X$, $d_w$ and $d_Y$ are
dimensions of $X$, $w$ and $Y$, respectively. We consider a supervised learning
setting, where a labeled training data-set $\mathcal{D}_{train}=\{ (X_u,Y_u)
\}_{u=1}^{N_{train}} $ is used to estimate classifier parameters $w$, where
$N_{train}$ are the total number of training samples. The training procedure is
the minimization of an appropriate loss function $\mathcal{L}: (w,\mathcal{D})
\rightarrow \mathbb{R}$ using an iterative algorithm like Stochastic Gradient
Descent. Therefore, the learned $w^*$ are the solution of the following
minimization problem $\mathcal{P}$.
\begin{align}
& w^* = \mathcal{P}(\mathcal{D}) = \argmin_w \mathcal{L}(w,\mathcal{D}) \label{eq:NN_problem}
\end{align}

Data augmentation involves applying an appropriate transformation $T: \mathcal{D}_{in} \rightarrow \mathcal{D}_{out}$ to a data-set (only $\mathcal{D}_{train}$ for our purposes) and hence expand it to an augmented data-set $T(\mathcal{D}_{train})$. We also use a validation data-set, $\mathcal{D}_{val}=\{ (X_u,Y_u) \}_{u=1}^{N_{val}}$ for cross-validation during training and a test data-set,  $\mathcal{D}_{test}=\{ (X_u,Y_u) \}_{u=1}^{N_{test}}$ for evaluating $g(X;w)$ post-training. The evaluation can be done using a suitable metric $\mathcal{M}: (w,\mathcal{D}) \rightarrow \mathbb{R}$ which maybe different from $\mathcal{L}$ above. Our aim is to find $T$ such that the estimated parameters $w_{aug}=\mathcal{P}(T(\mathcal{D}_{train}))$ perform better than $w_{train}=\mathcal{P}(\mathcal{D}_{train})$ in terms of the chosen metric, i.e. $\mathcal{M}(w_{aug},\mathcal{D}_{test})$ is more desirable than $\mathcal{M}(w_{train},\mathcal{D}_{test})$.

\subsection{Exploiting SAR Phenomenology for Data Augmentation}\label{teach}
Simple approaches to designing transformations $T$ for data augmentation consider translation invariance and symmetry of objects around its main axis, to introduce discrete pixel shifts and flips along the cross-range dimension. Our augmentation strategy goes further and use model-based transformations to improve the network's knowledge about two confounding factors namely pose, and scintillation effects due to shifts in range domain. Our method allows synthesizing new poses in a close neighborhood of existing poses based on a sparse modeling of the existing training data set, that exploits the spatial sparsity and the scattering centers' limited persistence. 

An overview of our approach is as follows: For every image in the training
dataset, We fit a sparse model in the phase-history domain exploiting SAR
phenomenology. This model is referred as the PH model henceforth. We utilize the
continuity of this PH model in azimuth domain to extrapolate phase-history
measurements and synthesize new images in a close neighborhood of the original
image.  The PH model also allows introducing arbitrary valued sub-pixel shifts
in both range and cross-range dimensions to images at both the original and
synthesized poses. These fractional shifts provide information to the
network regarding scintillation effects further improving its generalization
capability. In the following section we describe our modeling and pose synthesis
strategy in full detail


\subsection{Modeling and Pose Synthesis Methodology}\label{pose}

\begin{figure}[ht]
\centering
\includegraphics[trim=0 0 0
0,clip,width=0.4\textwidth]{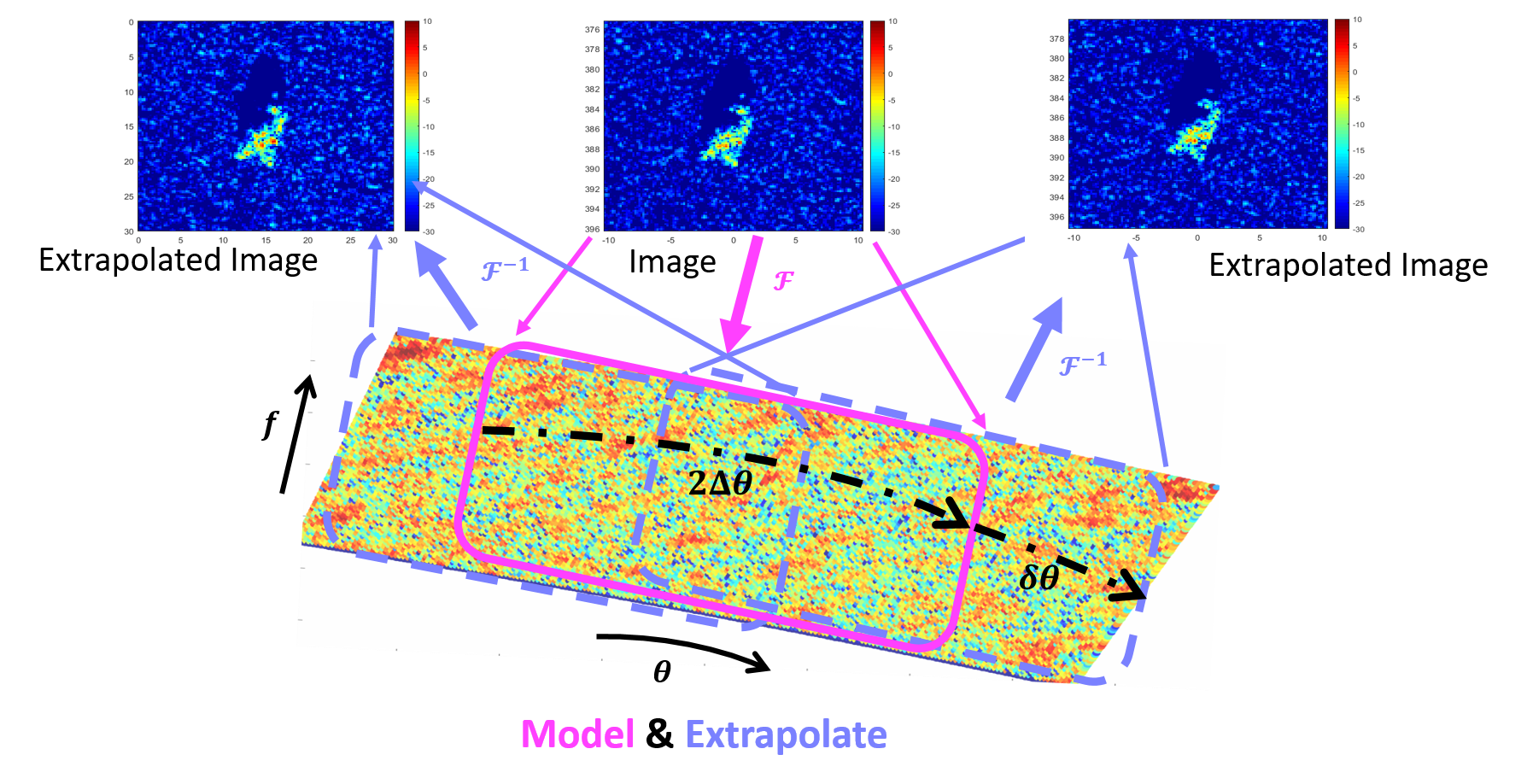}

\caption{Pose synthesis using the Phase History model. $\mathcal{F}$ denotes the Fourier Transform operator. The phase history collected over azimuth span of  $2\Delta \theta=3\degree$ is extrapolated by $\delta \theta$ using the model $\mathbf{S^*(\boldsymbol{\theta};\mathbf{f})}$.}
\label{fig:Extrapolation}
\end{figure}

This section describes the pose synthesis methodology used for data augmentation using the PH model. This work builds on our earlier work, which focused on modeling the scattering behavior of targets in monostatic and bistatic setup~\cite{sugavanam2017limited,sugavanam2018approximating,sugavanam2019bistatic}. We first construct a model for each image in the training data-set and locally extrapolate the measured images using the model. We assume a SAR sensor operating in the spotlight mode has been used to create the images (as in the case of the MSTAR data-set). The images are translated from the spatial domain to the Cartesian frequency domain using the steps described in~\cite{brito1999sar}. Subsequently, we convert the frequency measurements to the polar coordinates to obtain the phase-history measurements described in~\cite{Cetin_thesis_2001}.

We consider a square patch on the ground of side lengths $L =30 m$ centered around the target. From the Geometric theory of diffraction, we assume that a complex target can be decomposed into a sparse set of scattering centers. The scattering centers are assumed to be $K$ point targets, described using  $\{ (x_k,y_k),h_k(\theta,\phi) \}_{k=1}^K$ where $(x_k,y_k) \in [-\frac{L}{2},\frac{L}{2}] \times [-\frac{L}{2},\frac{L}{2}]$ are the spatial coordinates of the point targets, $\theta$ is the azimuthal angle, $\phi$ is the angle of elevation of the radar platform and $h_k(\theta,\phi)$ are the corresponding scattering coefficients that depend on the viewing angle. The samples of the received signal after the standard de-chirping procedure are given by
\begin{align}
& s(f_m;\theta,\phi) = \nn \\
&\sum_{k=1}^K h_k(\theta,\phi) \exp \pb{-j 4\pi
\frac{f_m\cos(\phi)}{c} \pb{x_k
\cos(\theta) + y_k \sin(\theta)} }, \label{eq:recSamples}
\end{align}

\nline
where $f_m$ are the illuminating frequencies such that $m \in [M]$, $M=\frac{2BL}{c}$, $B$ is bandwidth of transmitted pulse, $c$ is the speed of light and the notation $[M]$ denotes enumeration of natural numbers up till $M$. We estimate the function $h_k(\theta,\phi) \; \forall \; k \in [K]$ from the receiver samples. 

Parametric models for standard reflectors such as dihedral and trihedral were studied in
\cite{bistaticSAR_moses_2010,CornerReflectors_Sarbandi_1996,GTDModel_Potter_1995}. These models indicate that the reflectivity is a smooth function over the viewing angle, which is parameterized by the reflector's dimensions and orientation. Therefore, we exploit this smoothness to approximate this infinite-dimensional function using interpolation strategies
\cite{Rauhut2016_interpolation_weightedL1} with the available set of samples, $\Theta$, in the angle domain. We denote the sampled returns from the scene by the matrix $\mathbf{S} = NUFFT(X) \in \mathbb{C}^{N_{\theta} \times M}$, where NUFFT represents the non-uniform Fourier transform. The elements of $\mathbf{S}$ are defined as follows.
\begin{align}
 &s_{m,i} = n_{m,i}+ \sum_{k=1}^{K} h_k(\theta_i,\phi) \nn \\
 &\exp \pb{-j 4\pi
\frac{f_m\cos(\phi)}{c} \pb{x_k
\cos(\theta_i) + y_k \sin(\theta_i)}}.
\end{align}
where $n_{m,i}$ represents the measurement noise.
In order to solve the estimation problem, we assume that the
function $h_k$ has a representation in a basis set denoted by the matrix $\mathbf{\Psi} \in \mathbb{C}^{N_{\theta} \times D}$ of size $D$. For the MSTAR data-set, the elevation angles we work with are similar. We assume that the variation in $h_k$ with respect to $\phi$ is insignificant. This assumption leads to the following relation $h_k(\theta; \phi) = \sum_{v = 1}^{D} c_{v,k} \mathbf{\psi}_v(\theta) +  \epsilon_P$. The estimated phase-history matrix is now $\mathbf{\hat{S}}$ whose elements are given by

\begin{align}\label{eq:measureSAR}
 \hat{s}_{m,i} &= \hat{n}_{m,i} + \sum_{k=1}^{K}  \sum_{v = 1}^{D} c_{v,k} \psi_v(\theta_i)
\nn \\
&\exp
\pb{-j 4\pi
\frac{f_m\cos(\phi)}{c} \pb{x_k
\cos(\theta_i) + y_k \sin(\theta_i)} },
\end{align}

where $\hat{n}_{m,i}$ consists of the measurement noise and
the approximation error. To estimate the coefficients $c_{v,k}$ from the noisy measurements in \eqref{eq:measureSAR}, we discretize the scene with resolution of $\Delta R$ in $X,Y$ (range, cross-range) plane to get $K=N_R^2$ grid points, where $N_R = \frac{2BL}{c}$ is the number of range bins. Furthermore, we consider a smooth Gaussian function to perform the noisy interpolation. We partition the sub-aperture $2\Delta {\theta}$ into smaller intervals of equal length with a corresponding set containing the means of the intervals given by $\{ \hat{\theta}_v \}_{v=1}^{D}$, where $D=12$, which are used as the centroids for the Gaussian interpolating functions. We assume the width of the Gaussian function, $\sigma_G$ as a constant hyper-parameter whose selection is described in section \ref{DHP}. Hence, $\sigma_G$ is the constant minimum persistence of the scattering center in azimuth that we wish to detect. The radial basis functions used are

\begin{align}\label{eq:G_kernel}
\psi_v (\theta)=\exp\pb{-\pb{\frac{\theta -
\hat{\theta}_v}{2\sigma_G} }^2 }
\end{align}
The elements of $\mathbf{\hat{S}}$ due to scattering centers located at the discrete grid points are now given by

\begin{align}\label{eq:final_s_hat}
\hat{s}_{m,i} &=  \hat{n}_{m,i} + \sum_{k=1}^{N_R^2}\sum_{v=1}^{D}  c_{v,k} \mathbf{\psi}_v(
\theta_i) \nn \\
& \exp \pb{-j 4\pi
\frac{f_m\cos(\phi)}{c} \pb{x_k
\cos(\theta_i) + y_k \sin(\theta_i)} }.
\end{align}

Here, the discrete grids for $(x_k,y_k)$ and $(\theta_i,f_m)$ are both known. Let the vectors containing all corresponding grid points for $x_k, y_k, \theta_i, f_m$ be referred as $\mbf{x}, \mbf{y}, \boldsymbol{\theta}, \mbf{f}$ respectively. The problem now is to find the coefficients $c_{v,k}$ that minimizes the error between $\mathbf{\hat{S}}$ and $\mathbf{S}$. Let vector $\mathbf{c_k} = [c_{1,k} \cdots c_{D,k}]^T$. To recover the structured signal $\mathbf{h}=[h_1 \cdots h_{N_R^2}]$, which represents the scattering coefficient of a sparse scene that has a sparse representation in an underlying set of functions, we solve the following linear inverse problem using a sparse-group regularization on $\mathbf{c_k} \forall \; k \in [N_R^2]$.

\begin{align}\label{eq:DA_problem}
\min_{\mathbf{C}}\pb{\sum_{k=1}^{N_R^2}  \lambda \norm{\mathbf{c}_k}_{2} +\norm{\mathbf{S}-\mathbf{\hat{S}}}_F} \Longleftrightarrow \min_{\mathbf{C}} J(\mathbf{C},\sigma_G)
\end{align}

where $\mathbf{C}$ refers to the matrix $[\mathbf{c_1} \cdots \mathbf{c_{N_R^2}}]$, $\sigma_G$ is a constant hyper-parameter and $\norm{\cdot}_2$, $\norm{\cdot}_F$ refer to the $l^2$, Frobenius norms respectively.

The elements of the recovered model, $\mathbf{S^*(\boldsymbol{\theta};\mathbf{f})}$ are now 
\begin{align} \label{eq:F}
    &s^*_{i,m} = \sum_{k=1}^{N_R^2}\sum_{v=1}^{D}  c^*_{v,k} \mathbf{\psi}_v(
\theta_i) \nonumber \\ 
&\exp \pb{-j 4\pi
\frac{f_m\cos(\phi)}{c} \pb{x_k
\cos(\theta_i) + y_k \sin(\theta_i)} }
\end{align}
where $c^*_{v,k}$ are the recovered coefficients.
The phase-history measurements are converted back to the image using overlapping sub-apertures spanning $3$ degrees in the azimuth domain as shown in Fig.~\ref{fig:Extrapolation}. We apply the same Taylor window with zero-padding and translate it back to the Cartesian coordinates before applying the Fourier transform to generate the images to augment the data-set.

\section{Experiments}\label{Exp}
\subsection{MSTAR Dataset}\label{data}
The MSTAR data-set consists of 10 classes, i.e., tanks (T62, T72), armored vehicles (BRDM2, BMP2, BTR60, BTR70), a rocket launcher (2S1), an air defense unit (ZSU234), a military truck (ZIL131), and a bulldozer (D7). The radar platform used in constructing the MSTAR data-set acquires the measurements using $N_p =100$ pulses over an aperture of $3$ degrees. The phase-history measurements obtained in the receiver are converted to images using the sub-aperture based method described in~\cite{SandiaMethod_SAR_1994}. The motion-compensation steps is followed by the application of a Taylor window to control the side-lobes. The measurements are zero-padded to obtain an over-sampled image using the Fourier transform. The complete MSTAR data-set used in \cite{zhong2017enlightening} was highly imbalanced. We replace the data-set used in \cite{zhong2017enlightening} with a balanced subset, which is referred to as standard operating conditions considered in (\cite{chen2016target}, \cite{huang2017transfer}). We denote this subset as SOC MSTAR data-set henceforth.

Similar to the existing literature, we use the images at depression angle, $\phi
= 17 \degree$ for training while images at $\phi=15 \degree$ form the test set.
Similar to \cite{zhong2017enlightening}, we crop the images to $64 \times 64$
with the objects in the center. Note that we crop the images right before
feeding it to the ANN. We perform the modeling and augmentation steps on the
original images. Since our paper's objective is to investigate the effects of
data augmentation, we work with much smaller training data-sets by artificially
reducing the size of our data-set at $\phi = 17 \degree$. We exponentially
sub-sample by extracting only $\mathcal{R}$ ratio of samples from each class
where $\mathcal{R} \in \{ 2^{-5},2^{-4},2^{-3},2^{-2},2^{-1},2^{0} \}$. We
ensure that the extracted images are uniformly distributed over the $[0,2 \pi]$
azimuthal angle domain for each sub-sampling ratio. This sub-sampling strategy
is essential to ensure that the learning algorithm gets a complete view of the
vehicle's scattering behavior. We further select $15 \%$ of uniformly
distributed samples from this uniformly sub-sampled data as the validation set
and utilize the remaining $85 \%$ as the new training set. The training-data
$\mathcal{D}_{train}$ includes the flip augmentation along cross-range
domain~\cite{zhong2017enlightening} and real-time translations along both range and cross-range. These translations (in no. of pixels) are randomly sampled from the set $\{-6,-4,-2,0,2,4,6\}$, at every epoch and such a $\mathcal{D}_{train}$ is just referred to as baseline data henceforth.
We also include the flip-augmentation in the final validation set
$\mathcal{D}_{val}$ and no augmentations are included in the final test-set
$\mathcal{D}_{test}$. We form our $T(\mathcal{D}_{train})$ by performing the
proposed pose augmentation on each radar image in the training data-set as
described in the section~\ref{pose}. Additionally, our net transformation $T$ also includes
sub-pixel level translations as well as real-time pixel-level
translations~\cite{ding2016convolutional} in range and cross-range domain. We use sub-pixel shifts of $\frac{1}{2}$ pixel corresponding to approximately $0.15m$ displacement in the Y-direction (range) as well as X-direction (cross-range) of the scene, where each pixel corresponds to $0.3m$ in range and cross-range domain.

\begin{figure*}[htbp]
\centering
\includegraphics[width=\textwidth,height=6cm,keepaspectratio]{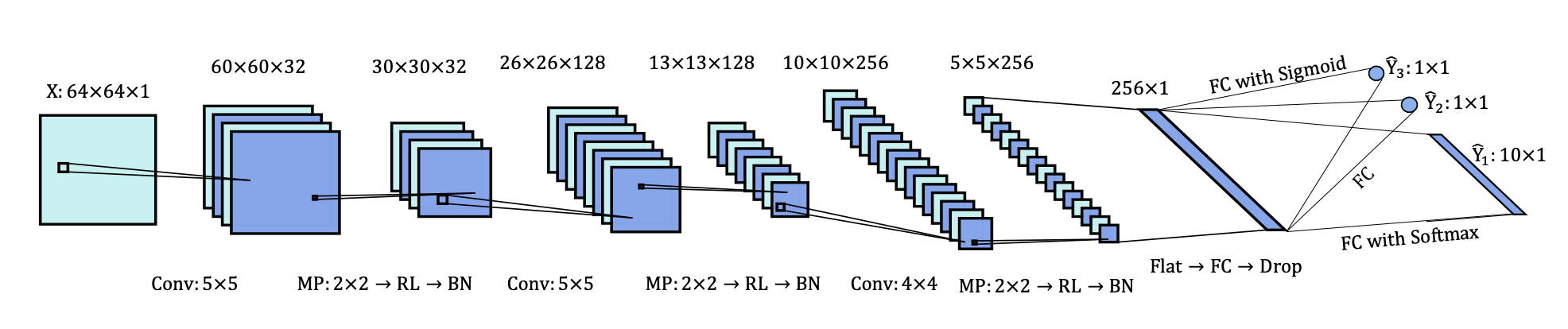}
\caption{The Neural Network Architecture. The abbreviations used are as follows. Conv is Convolutional Layer followed by the kernel height $\times$ width. MP is max-pooling followed by the pooling-size  as height $\times$ width. RL, BN, Flat, Drop and FC are ReLU, Batch-Normalization, Flattening, Dropout and Fully-Connected layers respectively. Sizes of feature maps are mentioned at their top as height $\times$ width $\times$ channels.}\label{fig:Network}
\end{figure*}

\subsection{Network Architecture}\label{network}
Our data augmentation algorithm is decoupled from the network architecture realizing the ATR algorithm. For the experiments we choose we utilize a state-of-the-art network architecture inspired by \cite{zhong2017enlightening} and shown in Fig.~\ref{fig:Network}. We modify the network and use batch-normalization layers after the ReLU activation in convolutional layers. We defer the use of dropout in convolutional layers since batch-normalization regularizes the optimization procedure~\cite{luo2018understanding}. After the last convolutional layer, we flatten out all the feature values and use a fully connected (FC) layer followed by a dropout layer. We further modify the cosine cost, used for pose-awareness in \cite{zhong2017enlightening}, to a pair of simpler costs using features $Y_2 = \sin( \theta )$ and $Y_3 = \mathds{1}_{A}( \theta )$ where $\theta$ is the azimuthal angle and $\mathds{1}_{A}$ is the indicator function over set $A= [\frac{-\pi}{2},\frac{\pi}{2}]$.  These two features uniquely determine the azimuthal angle and remove the need for a cosine distance loss. In our experiments, we found that this modification to the loss function resulted in improving the convergence of the optimization procedure while training the model. The loss function $\mathcal{L}$ to find the network parameters is now

\begin{align}
\hfill \mathcal{L}(w,\mathcal{D}) &= \bar{\E}_\mathcal{D} [ \mathcal{L}_1(w,X,Y_1)+\mathcal{L}_2(w,X,Y_2)+ \nonumber \\ \hfill &\mathcal{L}_3(w,X,Y_3)]\\
\hfill \mathcal{L}_1(w,X,Y_1) &= -\sum_{p=1}^{10} Y_{1,p} \log(\hat{Y}_{1,p}(w,\abs{X})) \nonumber\\
\hfill \mathcal{L}_2(w,X,Y_2)&= (Y_2- \hat{Y}_2(w,\abs{X}))^2 \nonumber\\
\hfill \mathcal{L}_3(w,X,Y_3)&= -Y_3 \log(\hat{Y}_3(w,\abs{X}))\nonumber \\  \hfill &-(1-Y_3) \log(1-\hat{Y}_3(w,\abs{X}))\nonumber
\end{align}

where $\abs{.}$ denotes the absolute value, $X \in \mathbb{C}^{64 \times 64},
Y_{1} \in \{ 0,1 \}^{10 \times 1}, Y_{2} \in [-1,1], Y_{3} \in \{ 0,1 \}$ refer
to complex radar images, the one-hot vector of the 10 classes, $\sin(\theta)$
and $\mathds{1}_{A}( \theta )$ respectively. $\bar{\E}_\mathcal{D}$ refers to
the empirical mean over data-set $\mathcal{D}$ and $Y_{1,p}$ is the $p^{th}$
component of the vector $Y_{1}$. All the quantities with $ \hat{} $ (hat) are
the corresponding estimates given by the ANN.

\begin{figure*}[ht]
\centering
\includegraphics[trim=0 0 0
0,clip,width=\textwidth,keepaspectratio]{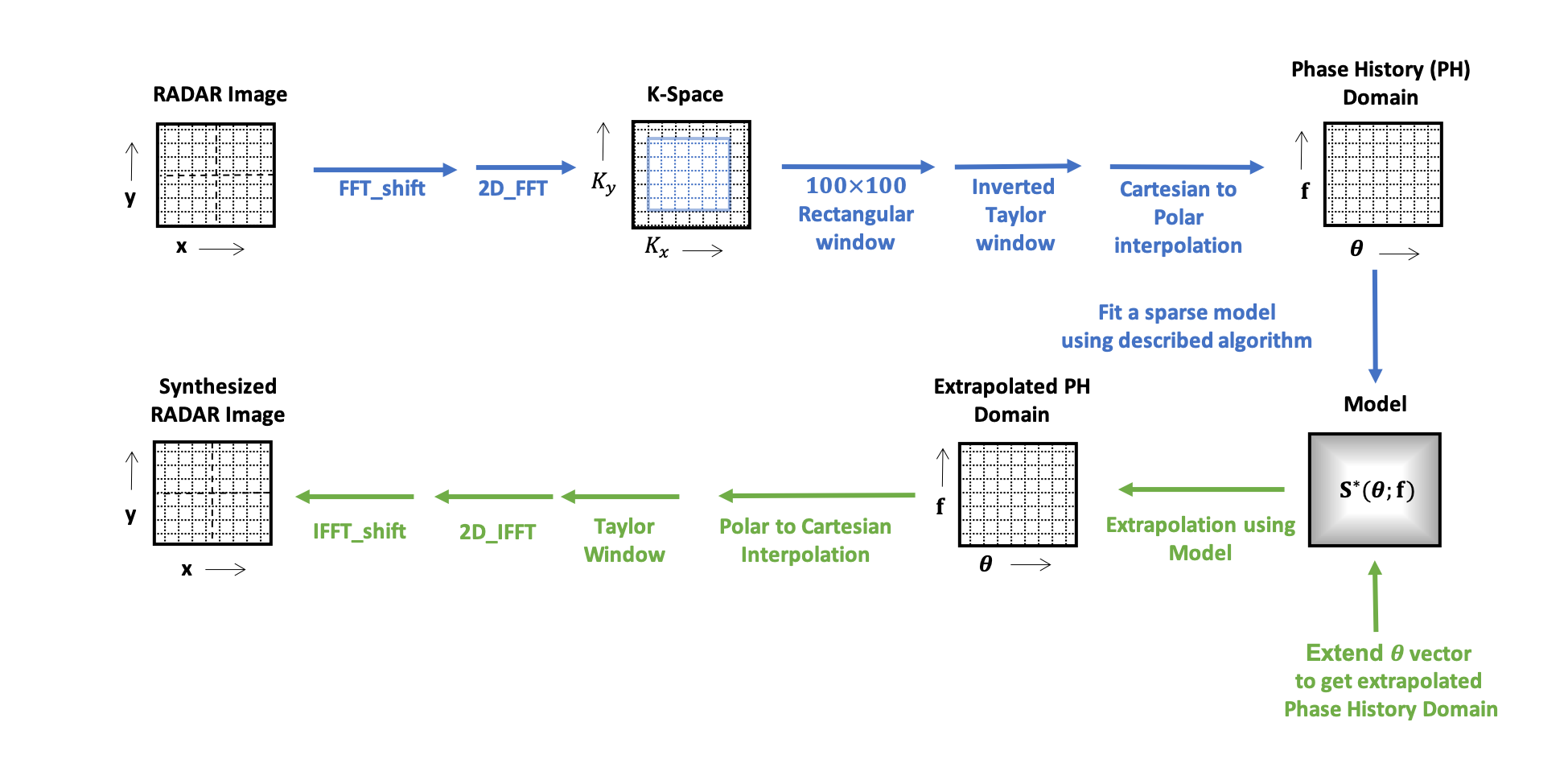}
\caption{Overview of the Image Synthesizing Procedure. All boxes with grid-lines represent matrices of complex values across an $(x_k,y_k) \; \forall \; k \in [N_R^2]$ grid where Cross-range and Range are labeled and along an $(\theta_i,f_m) \; \forall \; i \in [N_\theta],  m \in [M]$ grid where $\boldsymbol{\theta}$ and $\mbf{f}$ are labeled. Blue Arrows represent pre-model fitting stage and Green Arrows represent post-model fitting stage.}\label{fig:Img_Process}
\end{figure*}

\subsection{Experimental Setup}\label{setup}
The experiments were done using the network described in section \ref{network} on the data-sets described in section \ref{data}. This model is trained on a local machine with a Titan Xp GPU. The Tensorflow (1.10) \cite{abadi2016tensorflow} library is used for the implementation through its python API. We use the ReLU activation function everywhere except the final output layers of $\hat{Y}_1$, $\hat{Y}_2$, and $\hat{Y}_3$ where we use Softmax, Linear and Sigmoid activations respectively. 

An overview of the processing steps to synthesize radar images, starting from the complex radar data, is described in Fig. \ref{fig:Img_Process}. We first transform the image to K-space by inverting the transformations applied to the MSTAR data to get the phase history representation. Using the header  information from the MSTAR dataset, we determine the discrete grids for $(x_k,y_k)$ and $( \theta_i,f_m)$. Next, we estimate the model coefficients by solving the optimization problem described in equation \eqref{eq:DA_problem}. As a result, we obtain the model,  $\mathbf{S^*(\boldsymbol{\theta};\mathbf{f})}$, given by equation \eqref{eq:F}. This model is further used to synthesize new columns of phase-history data (or extending the $\boldsymbol{\theta}$ vector) and consequently produce a synthesized image by the procedure described in section \ref{pose} followed by transformation of phase history data to complex-valued image data. Complete MATLAB code to perform our proposed augmentations on MSTAR dataset is available at \url{https://github.com/SENSE-Lab-OSU/mstar_data_aug}.

We used the magnitude of the complex-valued radar data as input, $ X $, in agreement with the existing literature for training the network. We normalize all input images to the unit norm to reduce some undesired effects due to the Gaussian kernel in extrapolation. We also remove all the synthetic images at poses already in the corresponding training set. Then, the optimization problem in equation \eqref{eq:NN_problem} is solved using the off-the-shelf, Adam variant of the Mini-batch Stochastic Gradient Descent optimizer with a mini-batch size of $64$. The training is carried out for many epochs ($>400$), using the early-stopping criterion, and the model is saved for the best moving average validation performance metric. Although we care about accuracy (percentage of samples classified correctly) as the performance metric, the $\mathcal{D}_{val}$ here gets small, especially for small $\mathcal{R}$ values thereby saturating the validation accuracy at $100 \%$, yielding this metric less useful. Instead, we monitor the minimum classification loss, $\mathcal{L}_1$, as the validation performance metric. We report the percentage error (or misclassification), which is $100-accuracy$ as the test performance results.

\subsection{Determining Hyper-parameters}\label{DHP}
The PH model for each image and the neural network model introduce a set of hyper-parameters. We will now explain our choices of a subset of them and mention some others. The neural network's hyper-parameters are kept at the Tensorflow (1.10) library's default values unless specified.

The PH model has 2 main hyper-parameters, the $\sigma_G$ and $\delta\theta$.
We determine optimum $\sigma_G$ for every image by minimizing the following equation over all possible values of it, using a simple line-search.

$$
\sigma_G^* = \argmin_{\sigma_G} \sqb{\min_{\mathbf{C}} J(\mathbf{C},\sigma_G)}
$$

For determining appropriate $\delta\theta$, we choose the heuristic approach of grid search. We generate samples up to $6 \degree$ because the approximation error increases beyond that. We choose an appropriate $\delta\theta$ by running a grid search over a factor $\eta$ such that $\delta\theta=min\{6 \degree, \eta \sigma_G^*\}$. This is because the amount of possible extrapolation per image will depend on the corresponding kernel-width $\sigma_G^*$. We run the training on the smallest subset of the data-set at sub-sampling ratio of $2^{-5}$ for searching over a grid of 3 values, $\eta= \{1,2,3\}$.  We choose $\eta=3$ as it gives the best validation performance. Although we experimented with $\eta > 3$, we found the results comparable to $\eta = 3$. 

For the neural network model, we set the dropout rate for the last fully connected layer at 0.2.

\section{Results}\label{Res}

\begin{figure*}
\centering
\subcaptionbox{Radar image at a viewing angle of $\theta_c=57\degree$ generated by rotating the closest available in the sub-sampled dataset at $\theta_a=56\degree$ (from \cite{zhai2019sar}).\label{fig:naiveRotation}}
[.9\linewidth]{\includegraphics[width=0.72\textwidth,keepaspectratio]{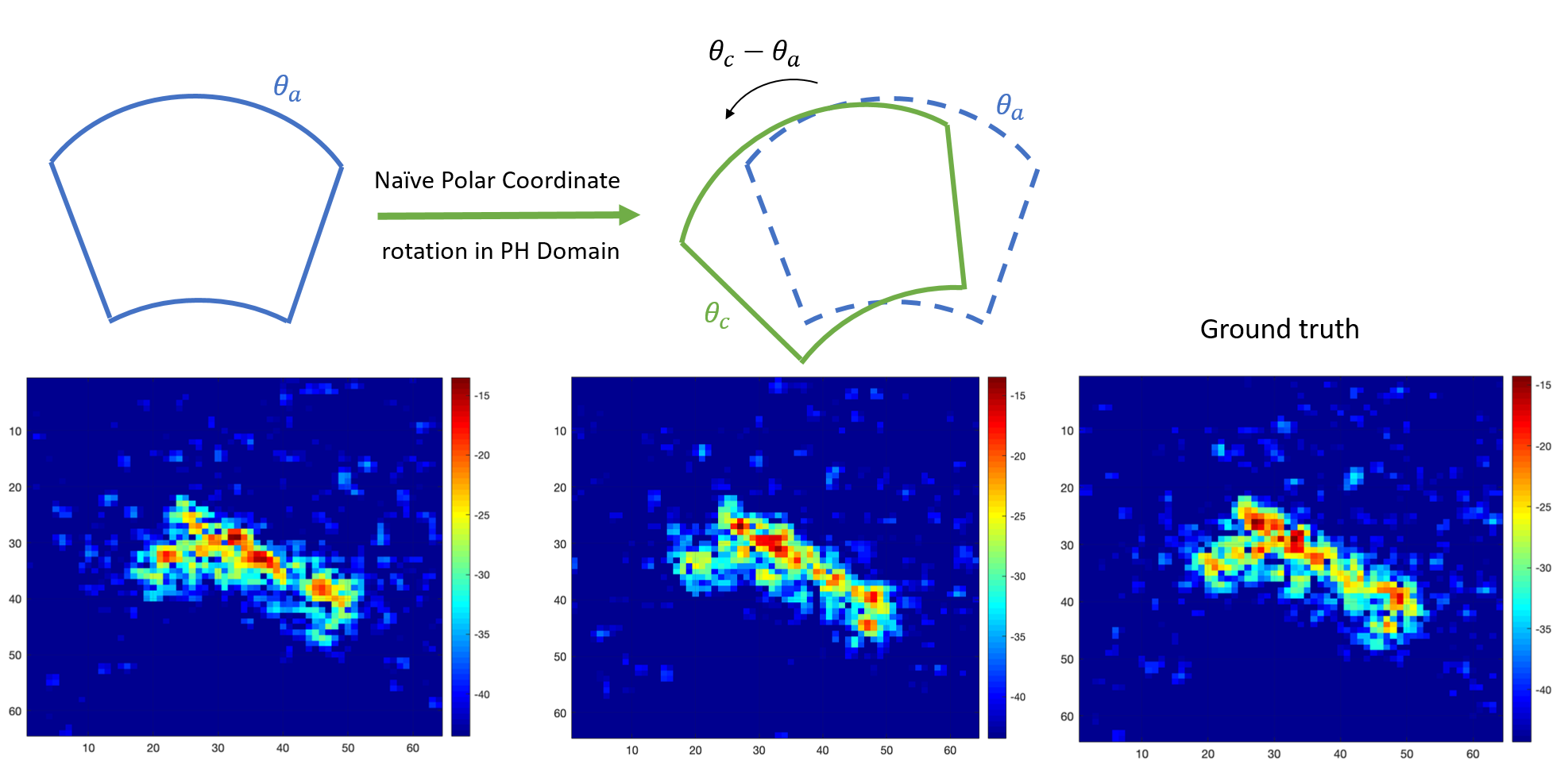}}
\subcaptionbox{Linear interpolation strategy proposed in \cite{ding2016convolutional}. Here, $\alpha$ and $\beta$ can be inferred from equation \ref{eq:Linear Interp} and the closest poses to $\theta_c=57\degree$ in the subsampled dataset were $\theta_a=56\degree$ and $\theta_b=85\degree$.\label{fig:linearInterp}}
[.9\linewidth]{\includegraphics[width=0.72\textwidth,keepaspectratio]{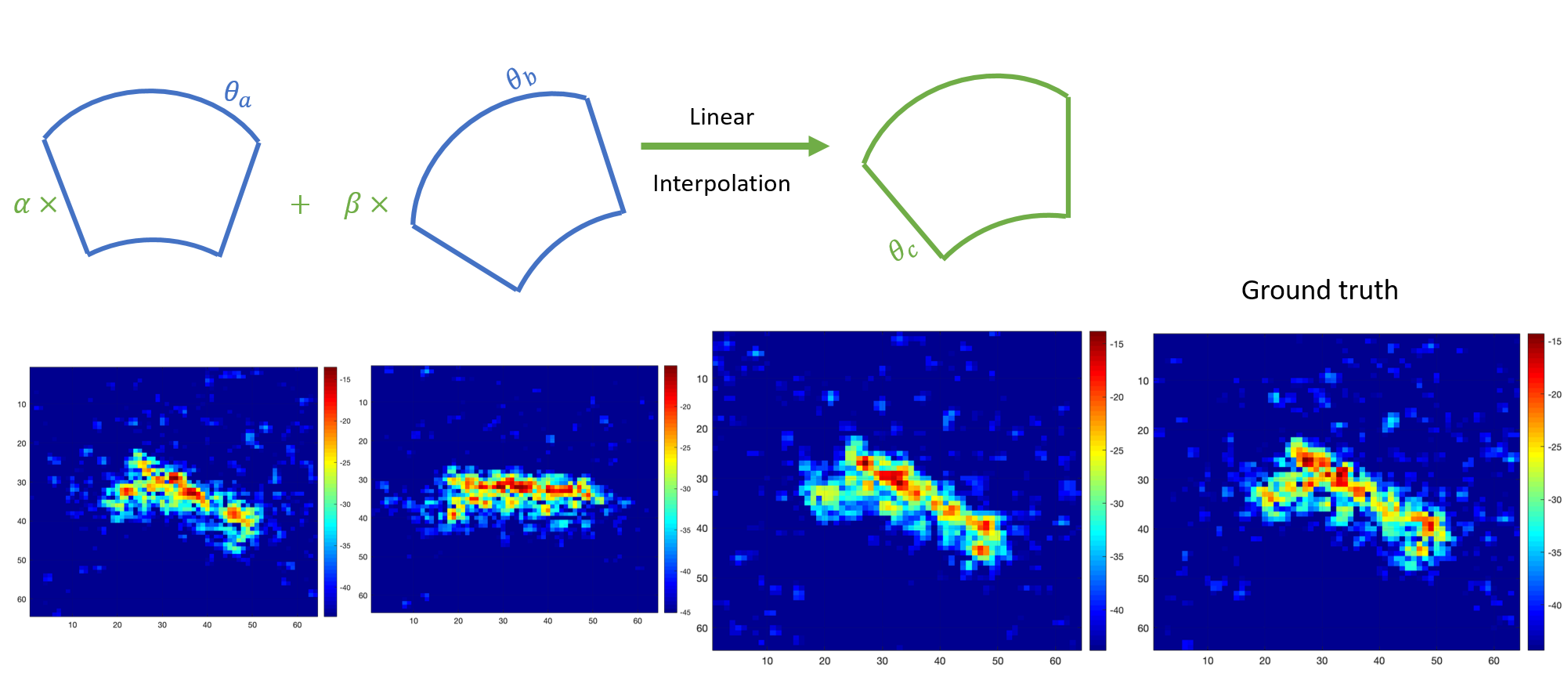}}\\
\subcaptionbox{Radar image at the azimuth angle of $\theta_a=56\degree$ is first approximated using the proposed set of basis functions in the frequency domain. Measurements from unobserved viewing angles are synthesized using the model in frequency domain, which is used to create the image at the viewing angle of $\theta_c=57\degree$.\label{fig:ProposedMethod}}
[.9\linewidth]{\includegraphics[width=0.72\textwidth,keepaspectratio]{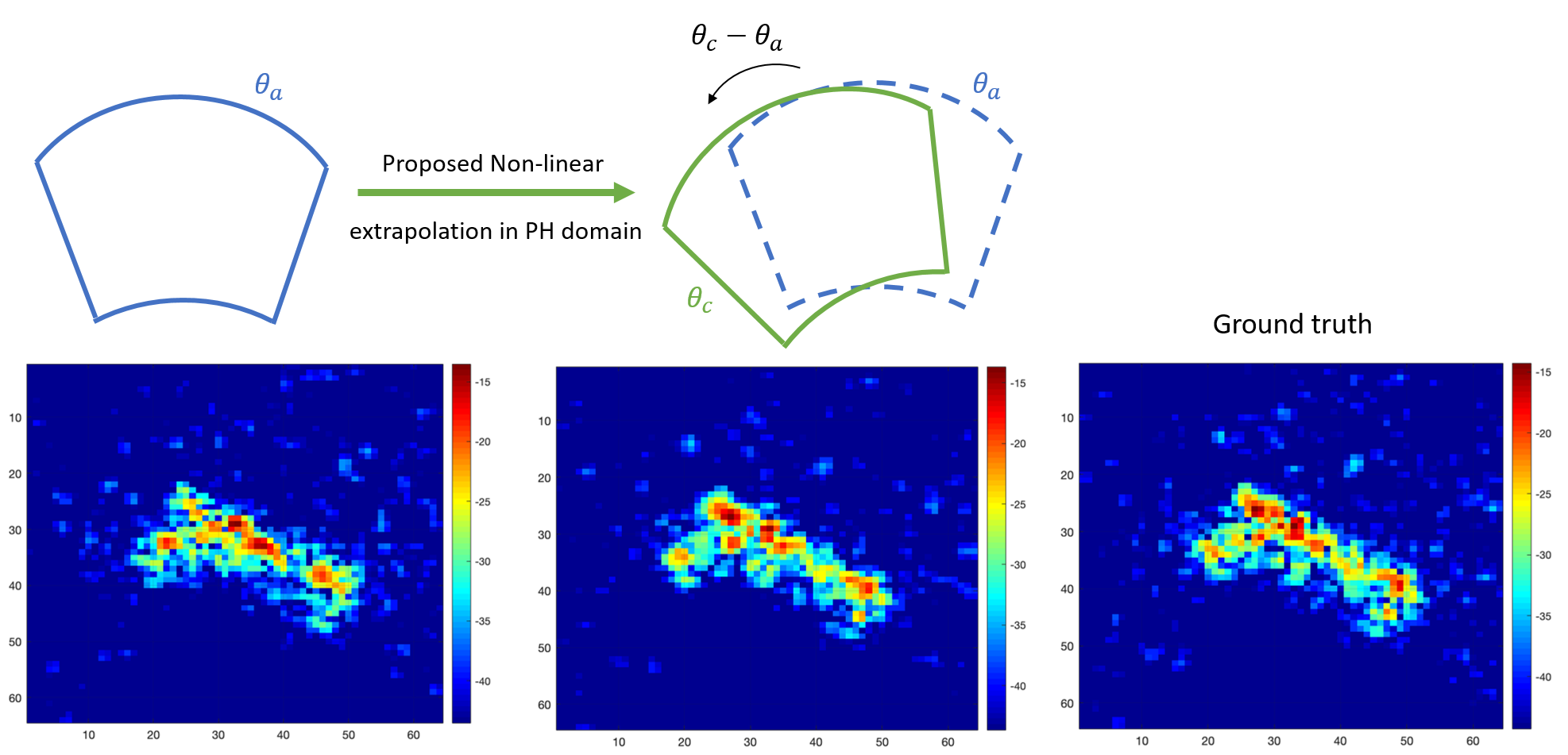}}\\
\caption{Comparison of radar images synthesized for class T62 at viewing angle $\theta_c=57\degree$ using different augmentation strategies}\label{fig:imagesT62}
\end{figure*}

\begin{table*}
	\begin{center}
	\caption{Test Errors corresponding to the analysis plot (Fig. \ref{fig:MainResults}). \label{table:MainResults}}
	\begin{tabular}{ | c | l | l | l | l |}
    \hline
	  \multicolumn{1}{|p{2cm}|}{\centering Sub-sampling Ratio ($\mathcal{R}$)}
	  & \multicolumn{1}{|p{2cm}|}{\centering Baseline Data (B)}
	  & \multicolumn{1}{|p{3cm}|}{\centering Adding our Sub-pixel shifts (B+S)}
	  & \multicolumn{1}{|p{3cm}|}{\centering Adding our poses and sub-pixel shifts (B+S+P)}
	  & \multicolumn{1}{|p{2cm}|}{\centering Full-data Features (F)} \\\hline
	  
    $2^{0}$ & 0.50 $\{B_0\}$ & 0.58 $\{S_0\}$ & 0.37 $\{SP_0\}$ & 0.50 $\{F_0\}$ \\\hline 
    $2^{-1}$ & 1.44 $\pm$ 0.33 $\{B_1\}$ & 1.03 $\pm$ 0.08 $\{S_1\}$ & 0.54 $\pm$ 0.00 $\{SP_1\}$ & 0.72 $\pm$ 0.06 $\{F_1\}$ \\\hline 
    $2^{-2}$ & 4.21 $\pm$ 0.83 $\{B_2\}$ & 2.33 $\pm$ 0.33 $\{S_2\}$ & 1.00 $\pm$ 0.34 $\{SP_2\}$ & 0.99 $\pm$ 0.17 $\{F_2\}$ \\\hline 
    $2^{-3}$ & 10.99 $\pm$ 0.73 $\{B_3\}$ & 5.68 $\pm$ 0.67 $\{S_3\}$ & 3.32 $\pm$ 1.31 $\{SP_3\}$ & 1.22 $\pm$ 0.26 $\{F_3\}$ \\\hline 
    $2^{-4}$ & 18.78 $\pm$ 2.40 $\{B_4\}$ & 14.02 $\pm$ 0.28 $\{S_4\}$ & 7.33 $\pm$ 0.54 $\{SP_4\}$ & 2.07 $\pm$ 0.21 $\{F_4\}$ \\\hline 
    $2^{-5}$ & 32.38 $\pm$ 2.93 $\{B_5\}$ & 29.98 $\pm$ 2.62 $\{S_5\}$ & 18.66 $\pm$ 3.22 $\{SP_5\}$ & 4.55 $\pm$ 0.57 $\{F_5\}$ \\\hline 
	\end{tabular}
	\end{center}
\end{table*}

\begin{table*}
	\begin{center}
	\caption{Test Errors corresponding to the comparative plot (Fig. \ref{fig:CompareResults}). \label{table:CompareResults}}
	\begin{tabular}{ | c | l | l | l | l |}
    \hline
	  \multicolumn{1}{|p{2cm}|}{\centering Sub-sampling Ratio ($\mathcal{R}$)}
	  & \multicolumn{1}{|p{2cm}|}{\centering Baseline Data (B)}
	  & \multicolumn{1}{|p{3cm}|}{\centering Augmenting with Naively Rotated Poses (B+R)}
	  & \multicolumn{1}{|p{3cm}|}{\centering Augmenting with Linearly Interpolated Poses (B+L)}
	  & \multicolumn{1}{|p{3cm}|}{\centering Augmenting with our poses and sub-pixel shifts (B+S+P)} \\\hline
	  
    $2^{0}$ & 0.50 $\{B_0\}$ & 0.62 $\{R_0\}$ & 0.50 $\{L_0\}$ & \textbf{0.37} $\{SP_0\}$ \\\hline 
    $2^{-1}$ & 1.44 $\pm$ 0.33 $\{B_1\}$ & 1.22 $\pm$ 0.19 $\{R_1\}$ & 1.22 $\pm$ 0.14 $\{L_1\}$ & \textbf{0.54} $\pm$ 0.00 $\{SP_1\}$ \\\hline 
    $2^{-2}$ & 4.21 $\pm$ 0.83 $\{B_2\}$ & 4.38 $\pm$ 0.72 $\{R_2\}$ & 2.56 $\pm$ 0.21 $\{L_2\}$ & \textbf{1.00} $\pm$ 0.34 $\{SP_2\}$ \\\hline 
    $2^{-3}$ & 10.99 $\pm$ 0.73 $\{B_3\}$ & 10.01 $\pm$ 1.72 $\{R_3\}$ & 7.26 $\pm$ 2.56 $\{L_3\}$ & \textbf{3.32} $\pm$ 1.31 $\{SP_3\}$ \\\hline 
    $2^{-4}$ & 18.78 $\pm$ 2.40 $\{B_4\}$ & 19.83 $\pm$ 2.21 $\{R_4\}$ & 12.99 $\pm$ 1.10 $\{L_4\}$ & \textbf{7.33} $\pm$ 0.54 $\{SP_4\}$ \\\hline 
    $2^{-5}$ & 32.38 $\pm$ 2.93 $\{B_5\}$ & 32.05 $\pm$ 7.58 $\{R_5\}$ & 30.79 $\pm$ 3.04 $\{L_5\}$ & \textbf{18.66} $\pm$ 3.22 $\{SP_5\}$ \\\hline 
	\end{tabular}
	\end{center}
\end{table*}

\begin{table*}
	\begin{center}
	\caption{Results from various SAR-ATR efforts using SOC MSTAR data-set.\label{table:SOTA}}
	\begin{tabular}{ | c | c | c | }
    \hline
      Method & Error(\%) using $100\%$ data & Error(\%) using $\leq 20\%$ data \\\hline
	  SVM (2016) \cite{song2016sparse} & 13.27 & 47.75 (at $20\%$)\\\hline
	  SRC (2016) \cite{song2016sparse} & 10.24 & 36.35 (at $20\%$)\\\hline
	  A-ConvNet (2016) \cite{chen2016target} & 0.87 & 35.90 (at $20\%$)\\\hline
	  Ensemble DCHUN (2017) \cite{lin2017deep} & 0.91 & 25.94 (at $20\%$)\\\hline
	  CNN-TL-bypass (2017) \cite{huang2017transfer} & 0.91 & 2.85 (at $18\%$) \\\hline
	  ResNet (2018) \cite{fu2018small} & 0.33 & 5.70 (at $20\%$)\\\hline
	  DFFN (2019) \cite{yu2019high} & \textbf{0.17} & 7.71 (at $20\%$)\\\hline
	  Our Method & 0.37 & \textbf{1.53} (at $18\%$) \\\hline
	\end{tabular}
	\end{center}
\end{table*}

\begin{table*}
	\begin{center}
	\caption{Confusion matrix for classifier corresponding to $\mathcal{R}=2^{-4}$ with no augmentation.\label{table:CM1}}
	\begin{tabular}{ |c | c | c| c | c | c| c | c | c| c | c | c|}
    \hline			
	  Class & 2S1 & BMP2 & BRDM2  &BTR60  & BTR70  &D7 &T62 &T72 &ZIL131  &ZSU234 & Error (\%) \\ \hline
        2S1 & 196 & 0 & 1 & 0 & 3 & 1 & 40 & 5 & 18 & 10 & 28.467 \\ \hline
        BMP2 & 21 & 117 & 2 & 18 & 10 & 0 & 1 & 23 & 3 & 0 & 40.0 \\ \hline
        BRDM2 & 9 & 1 & \textbf{256} & 1 & 0 & 1 & 0 & 0 & 6 & 0 & 6.569 \\ \hline
        BTR60 & 2 & 2 & 4 & 161 & 10 & 3 & 2 & 4 & 4 & 3 & 17.436 \\ \hline
        BTR70 & 21 & 13 & 1 & 23 & 130 & 1 & 0 & 6 & 0 & 1 & 33.673 \\ \hline
        D7 & 0 & 0 & 0 & 0 & 0 & \textbf{264} & 1 & 0 & 7 & 2 & 3.65 \\ \hline
        T62 & 5 & 0 & 0 & 2 & 0 & 1 & 234 & 4 & 22 & 5 & 14.286 \\ \hline
        T72 & 5 & 2 & 0 & 3 & 0 & 1 & 16 & 164 & 5 & 0 & 16.327 \\ \hline
        ZIL131 & 1 & 0 & 0 & 0 & 0 & 34 & 4 & 0 & 234 & 1 & 14.599 \\ \hline
        ZSU234 & 0 & 0 & 0 & 0 & 0 & 17 & 11 & 0 & 29 & 217 & 20.803
    
    \\\hline
    \multicolumn{11}{|c|}{Overall} & 18.639
    \\\hline
	\end{tabular}
	\end{center}
\end{table*}
	
\begin{table*}
	\begin{center}
	\caption{Confusion matrix for classifier corresponding to $\mathcal{R}=2^{-4}$ with augmentation.\label{table:CM2}}
	\begin{tabular}{ |c | c | c| c | c | c| c | c | c| c | c | c|}
    \hline			
	  Class & 2S1 & BMP2 & BRDM2  &BTR60  & BTR70  &D7 &T62 &T72 &ZIL131  &ZSU234 & Error (\%)\\ \hline
        2S1 & \textbf{251} & 0 & 0 & 1 & 0 & 1 & 9 & 8 & 4 & 0 & 8.394 \\ \hline
        BMP2 & 4 & \textbf{169} & 0 & 4 & 0 & 0 & 4 & 12 & 1 & 1 & 13.333 \\ \hline
        BRDM2 & 16 & 8 & 243 & 0 & 0 & 0 & 0 & 0 & 6 & 1 & 11.314 \\ \hline
        BTR60 & 2 & 1 & 4 & \textbf{172} & 6 & 1 & 3 & 1 & 1 & 4 & 11.795 \\ \hline
        BTR70 & 7 & 2 & 1 & 0 & \textbf{184} & 0 & 0 & 2 & 0 & 0 & 6.122 \\ \hline
        D7 & 0 & 0 & 0 & 0 & 0 & 263 & 0 & 0 & 0 & 11 & 4.015 \\ \hline
        T62 & 6 & 0 & 0 & 4 & 0 & 1 & \textbf{257} & 4 & 1 & 0 & 5.861 \\ \hline
        T72 & 1 & 0 & 0 & 0 & 0 & 0 & 10 & \textbf{183} & 0 & 2 & 6.633 \\ \hline
        ZIL131 & 6 & 0 & 0 & 0 & 0 & 8 & 6 & 1 & \textbf{244} & 9 & 10.949 \\ \hline
        ZSU234 & 0 & 0 & 0 & 0 & 0 & 0 & 2 & 0 & 0 & \textbf{272} & 0.73
    
    \\\hline
    \multicolumn{11}{|c|}{Overall} & \textbf{7.711}
    \\\hline
	\end{tabular}
	\end{center}
\end{table*}

\begin{figure*}
\centering
\subcaptionbox{Ablation Study.\label{fig:MainResults}}
[.5\linewidth]{\includegraphics[width=0.49\textwidth,keepaspectratio]{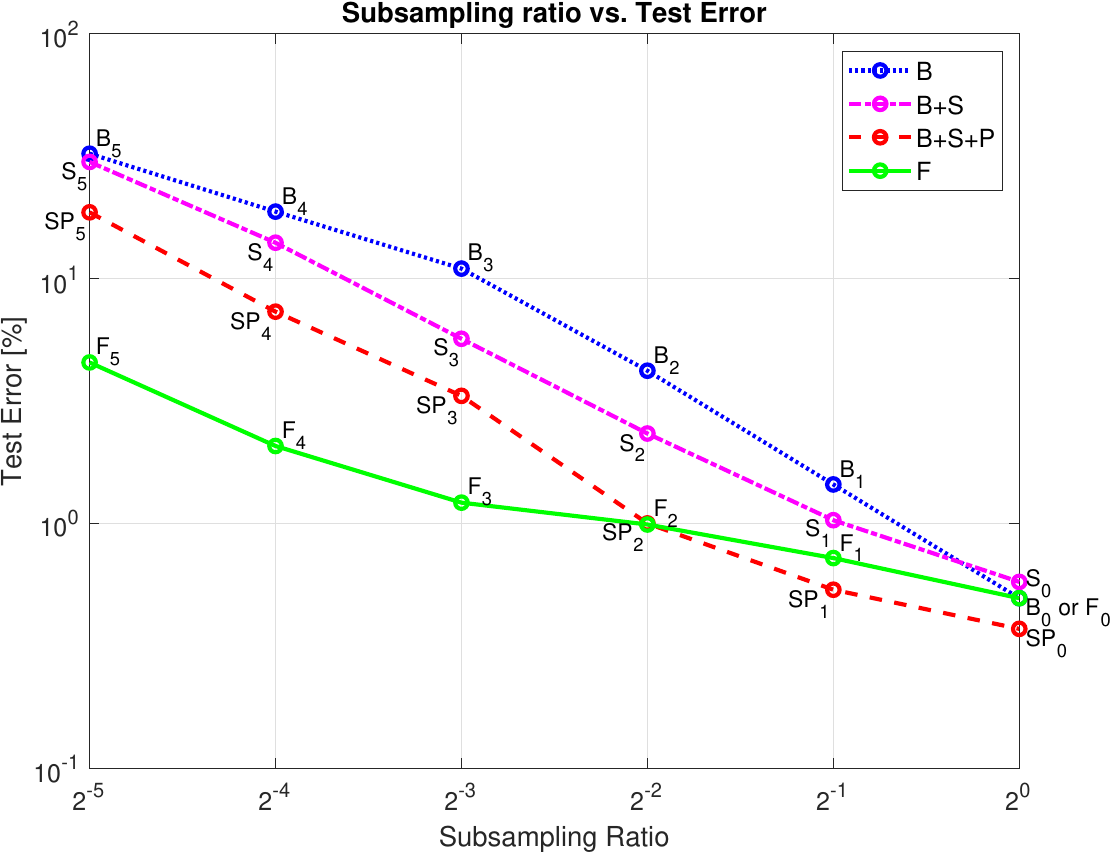}}%
\subcaptionbox{Comparison with other augmentations.\label{fig:CompareResults}}
[.5\linewidth]{\includegraphics[width=0.49\textwidth,keepaspectratio]{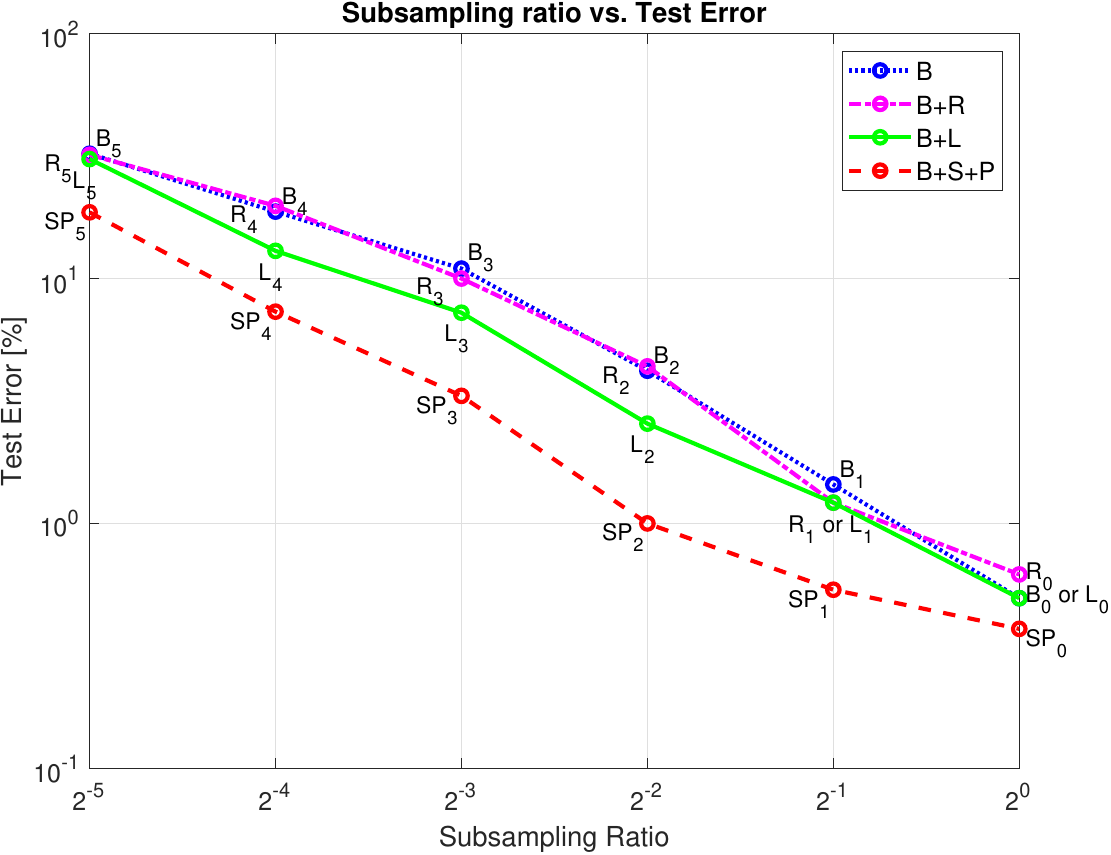}}
\caption{Quantitative Evaluation of the Proposed Approach}\label{fig:Results}
\end{figure*}

Scarcity of training data affect the performance of the resulting  ATR classifier in two distinct ways: First, a small training data-set interferes with the ability to learn to extract informative features from the data. Second, given a set of features, limited training data results in suboptimal decision boundaries leading to poor generalization performance. We hypothesize that  data augmentation techniques primarily  improves the former effect, i.e. it improves test performance through enhanced training of the CNN's convolutional layers that serve as the feature extractors.
Our empirical results  presented below support this observation. With an adequate feature set, the classifier can be trained even with a small training data-sets and will generalize well. To disentangle the two effects, the convolutional layers of the network are trained with the augmented training data-set and the classifier layers (after and including the first FC layer) are trained using the corresponding non-augmented training data-set. 


For all the approaches we generate results for $6$ values of $\mathcal{R}$ corresponding to different sub-sampling ratios of the original training dataset.
All the models have the same architecture as described in section \ref{network} and use the same $\mathcal{D}_{test}$. The difference among them is the $\mathcal{D}_{train}$ and $\mathcal{D}_{val}$ used as described in section \ref{data}. For consistency of results, we repeat the process described in section \ref{data} to get four different $\mathcal{D}_{train}$ and $\mathcal{D}_{val}$ for each $\mathcal{R}$ (except for $\mathcal{R}=2^{-1}$ and  $\mathcal{R}=2^0$ where only two and one such unique data-sets were possible, respectively) and report mean and standard deviation of the classification performance. The overall augmentation performance is summarized in tables \ref{table:CompareResults}, \ref{table:MainResults} and visualized in Fig. \ref{fig:Results}. 

\subsection{Ablation Study of the Proposed Approach}\label{ablation}

We perform an ablation study of the two proposed augmentations, sub-pixel and pose augmentations by incrementally adding them to the Baseline data.
We abbreviate data-sets as: Baseline data as B (that includes image domain flips and integer pixel translations), Baseline data with proposed sub-pixel augmentations as B+S, Baseline data with proposed sub-pixel and pose augmentations as B+S+P.
Moreover, to provide a lower bound on test-error of data augmentation approaches at all the sub-sampling ratios, we use a genie aided approach (non realizable in practice) by utilizing the full SOC data-set to learn CNN features, but use only the sub-sampled training data-set for the fully connected classifier layers. This forms the test-error curve referred as F (for Full-data) in Fig. \ref{fig:MainResults}.

The Full-data plot in the Fig. \ref{fig:MainResults} (values in table \ref{table:MainResults}), shows the importance of extracting good quality features, i.e., if we had access to all the poses, we would learn very good features. Having good features makes classification quite easy, evident from the low test errors even in very low data availability for learning the classifier. The sub-sampling has little effect on the generalization performance  for the genie-aided case. The Baseline data plot shows a considerable amount of test error, especially in low training-data availability. This test error is reduced in B+S data plot and further reduced in B+S+P data plot, which shows the effectiveness of both our strategies in improving the quality of features extracted by the CNN. Note that the majority of the improvement comes from the pose augmentations. For $\mathcal{R}=2^{-5}$, the proposed augmentations reduce the test error by more than 42\% as compared to the baseline approach that includes image domain flips and integer pixel translations. For $\mathcal{R}>2^{-2}$, the model using both proposed augmentations gives even better performance than genie-aided features. This makes sense because our augmentation strategy is able to fill-in pose information gaps successfully in the complete SOC data. However, there exists a considerable gap between Full-data and B+S+P data plots in the smaller data regimes, $\mathcal{R}<2^{-2}$. So, there still exists room for further improvements in aiding the network learning informative features for the data starved regimes.

The confusion matrices for a sample data-set at $\mathcal{R}=2^{-4}$ are shown in tables \ref{table:CM1} and \ref{table:CM2}. These tables clearly show that the performance has considerably improved, with the proposed augmentation of training data in low data availability. Not only that, but the performance has also improved over all classes except two. As there were four distinct sub-sampled data-sets at $\mathcal{R}=2^{-4}$, we picked the one which was a good representative of the average performance.

\subsection{Comparison with existing SAR-ATR models}\label{Compare_SAR_ATR}

Comparing our test error using the full SOC MSTAR data-set, it can be seen from table \ref{table:SOTA} that our approach is at par with the existing approaches when using all the data. We are interested in training the CNN models when data availability is extremely low, say $\le 60$ samples per class. To compare results from our approach to recent works in extremely low data-regimes, we utilized some results from \cite{yu2019high} and \cite{huang2017transfer}. We also conducted B+S+P experiments for $18\%$ data per class. These are also tabulated in table \ref{table:SOTA}. It is this extreme sub-sampling regime where our approach out-performs all other existing approaches. The proposed algorithm reduces the test error by
more than 46\% as compared to the next best approach of CNN-TL-bypass \cite{huang2017transfer}. We get the lowest test-error even when using the smallest portion of the data. We reiterate that most of the tabulated approaches are decoupled from our data augmentation approach. So, in principle, it may be possible to combine our data augmentation strategy with the existing approaches to get even better results.

Except for works \cite{zhai2019sar} and \cite{ding2016convolutional}, we did not find reproducible data augmentation strategies that explicitly synthesize samples at new poses. As pointed out earlier, our approach can be used in conjunction with most of the other strategies outlined in section \ref{RW}. So, we do a detailed comparison of our pose synthesis approach and sub-pixel level translations with pose synthesis methods in \cite{zhai2019sar} and \cite{ding2016convolutional}. We add real-time pixel level translations for all experiments. For the sake of completion, the simple rotations are produced in \cite{zhai2019sar} using the rotation matrix followed by appropriate cropping and the linearly interpolated poses are synthesized in \cite{ding2016convolutional} using the following equation:

\begin{align}\label{eq:Linear Interp}
I_{\theta_c}={CR}_{\theta_c}\pb{\frac{|\theta_b-\theta_c|R_{\theta_a}(I_{\theta_a})+|\theta_a-\theta_c|R_{\theta_b}(I_{\theta_b})}{|\theta_a-\theta_c|+|\theta_b-\theta_c|}}
\end{align}
where ${R}_{\theta}(I)$ denotes the rotation of radar image $I$ by $\theta$ degrees clockwise, ${CR}_{\theta}(I)$ denotes the same but counter-clockwise, $I_\theta$ denotes radar-image at the pose $\theta$, $\theta_c$ is the desired new pose, $\theta_a$ and $\theta_b$ are the poses closest to $\theta_c$ in the training data.

For a qualitative evaluation, we illustrate the images synthesized of the $T62$ tank for $\theta_c = 57\degree$ and the corresponding ground-truth image (which is part of the Full SOC data) is used for comparison purposes only. We observe in Fig.~\ref{fig:linearInterp} the synthesized image $\theta_c = 57\degree$ using $\theta_a = 56\degree$ and $\theta_b= 85 \degree$ using a sub-sampled data-set with $\mathcal{R}=2^{-4}$. We note that the dominant scattering centers in the synthesized image is different compared to the ground-truth. Next, we used the proposed model estimated for azimuth angle $56\degree$. It's evident from Fig.~\ref{fig:ProposedMethod} that the synthesized image from our method captures all the dominant scattering centers present in the ground truth. Finally, we use the rotation operator to synthesize $\theta_c = 57\degree$ using $\theta=56 \degree$. We observe that even the rotation is small the ground-truth image has a different scattering behavior, which does not get captured by rotation. 

For the quantitative evaluation, we compare the ATR performance at all sub-sampling ratios similar to previous section \ref{ablation}. We abbreviate data-sets as: Baseline data as B,
baseline data with simple rotations added (from \cite{zhai2019sar}) as B+R, baseline data with linearly interpolated poses (from \cite{ding2016convolutional}) as B+L and baseline data with proposed sub-pixel and pose augmentations as B+S+P.
 
The comparison is tabulated in table \ref{table:CompareResults} and can be seen in Fig. \ref{fig:CompareResults}. It is evident that our approach is significantly better than both, simple rotations and linearly interpolated poses, for this CNN based ATR task. For $\mathcal{R}=2^{-5}$, the proposed augmentations reduce the test error by more than 39\% as compared to the next best augmentation approach of \cite{ding2016convolutional}.

\section{Conclusion and Future Directions}\label{Conc}

In this paper, we proposed incorporating domain knowledge of SAR phenomenology into a CNN by way of data-augmentation. We presented a model based approach to  data-augmentation for training the neural network architecture to solve the ATR problem with limited labeled data. Through extensive simulation studies we showed the effectiveness the augmentation strategies by training a neural network with the augmented data-set synthesized from  phase-history models extracted from each available training image. Our results show that the proposed data augmentation strategy gave a significant improvement in the model's generalization performance compared to the baseline performance over a wide range of sub-sampling ratios. As presented, the phase-history approximation method is only valid in a local neighborhood of the  given azimuth angle. Future work could focus on fitting a single global model to every class jointly derived from all the training images. Such a global model can produce a diverse set of SAR images over larger pose variations. Since typically target image chips are not perfectly registered and aligned across different azimuth angles, the global model fit should incorporate unknown phase and spatial shifts for each image. As part of future research, we propose developing a network architecture to learn a unified model that can account for these phase errors and synthesize a larger data-set to improve the classifier's performance further.

\section*{Acknowledgements}

This research was partially supported by Army Research Office grant W911NF-11-1-0391 and NSF Grant IIS-1231577.

\bibliographystyle{IEEEtran}
\bibliography{references.bib}

\end{document}